\documentclass{article}
\usepackage[margin=2.5cm]{geometry}

\usepackage{epfml}

\usepackage{soul}
\usepackage{url}
\usepackage[utf8]{inputenc}
\usepackage[small]{caption}
\usepackage{graphicx}
\usepackage{amsmath}
\usepackage{booktabs}
\usepackage{lipsum}  
\usepackage{multirow}
\usepackage{booktabs}
\usepackage{siunitx}
\usepackage{mathtools}
\usepackage{subcaption}
\usepackage{xspace}
\usepackage{siunitx}

\usepackage{stmaryrd} % for arrow symbols
\makeatletter
\newcommand{\oset}[3][0ex]{%
  \mathrel{\mathop{#3}\limits^{
    \vbox to#1{\kern-1\ex@
    \hbox{$\scriptstyle#2$}\vss}}}}
\makeatother
\newcommand{\fedge}[1]{\oset{\shortrightarrow}{e}_{\hspace{-0.3mm}#1}}
\newcommand{\bedge}[1]{\oset{\shortleftarrow}{e}_{\hspace{-0.3mm}#1}}

\usepackage{xcolor}
\definecolor{myBlue}{rgb}{.204,.596,.859}

\definecolor{myRed}{rgb}{.906, .298, .235}

\newcommand{\idx}{\mathcal{I}}
\newcommand*{\defeq}{\stackrel{\text{def}}{=}}
\newcommand{\alg}{\textsc{Gretel}\xspace}
\providecommand{\bphi}{{\phi}}

\renewcommand{\paragraph}[1]{\vspace{2mm}\textbf{#1}}

\title{Extrapolating paths with graph neural networks}

\author{
Jean-Baptiste Cordonnier\footnote{Corresponding Author} ~ and ~ Andreas Loukas \vspace{3mm} \\ \vspace{1mm} 
\'{E}cole Polytechnique F\'{e}d\'{e}rale de Lausanne \\
\{jean-baptiste.cordonnier, andreas.loukas\}@epfl.ch
}

\date{}

\begin{document}

\maketitle

\begin{abstract}
We consider the problem of path inference: given a path prefix, i.e., a partially observed sequence of nodes in a graph, we want to predict which nodes are in the missing suffix. In particular, we focus on natural paths occurring as a by-product of the interaction of an agent with a network---a driver on the transportation network, an information seeker in Wikipedia, or a client in an online shop. Our interest is sparked by the realization that, in contrast to shortest-path problems, natural paths are usually not optimal in any graph-theoretic sense, but might still follow predictable patterns. 
 
Our main contribution is a graph neural network called \alg. Conditioned on a path prefix, this network can efficiently extrapolate path suffixes, evaluate path likelihood, and sample from the future path distribution. Our experiments with GPS traces on a road network and user-navigation paths in Wikipedia confirm that \alg is able to adapt to graphs with very different properties, while also comparing favorably to previous solutions. 
\end{abstract}

\section{Introduction}
\label{sec:intro}

Can a graph neural network learn to extrapolate paths from examples? 
Rather than attempting to connect nodes based on some graph-theoretic objective function (e.g., by looking for a shortest path), this work focuses on {naturally occurring paths}. 
Such paths appear whenever an agent tries to reach a target by moving between adjacent nodes in a graph. 
The agent for example may be a driver that is navigating through a road network or a knowledge seeker browsing through Wikipedia articles. 
Given the graph and a partial knowledge of the path our goal is to predict the future position of the agent.

Path inference problems are demanding because natural paths tend to differ qualitatively from shortest paths.
The choice of the agent at every step may depend on a number of factors, such as the structural properties of the graph and the conceptual similarity of nodes as perceived by the agent. For instance, when looking for information in Wikipedia it has been observed that seekers' decisions are correlated with their perception of article similarity and degree~\cite{wikispeedia}.
At the same time, some form of directionality is involved in path formation, in the sense that an agent's actions can be conditioned on the entire history of its trajectory. Making a parallel to Euclidean domains, a path can be thought as `straight' when the agent moves towards nodes that are far from where it was in the past and `circular' when it returns to its starting position. Contrasting our geometric intuition however, the space here is non-Euclidean and what is far or close should be determined in light of the graph structure.

\begin{figure*}[t!]
\centering{
\hspace{-1.0mm}\resizebox{83mm}{!}{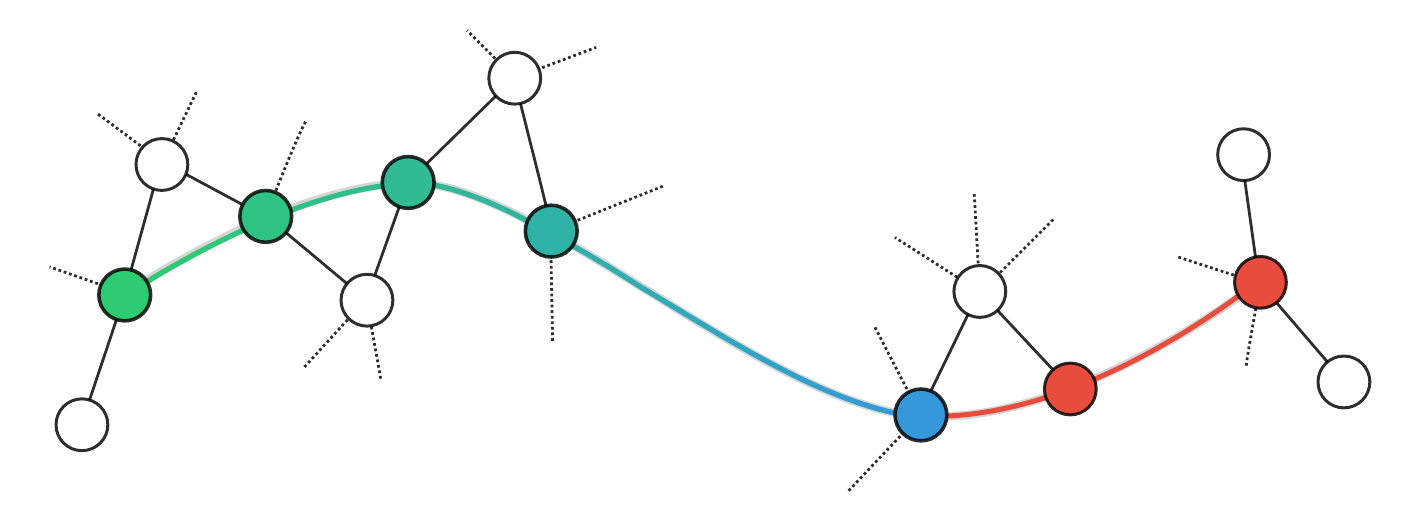} 
}
\hfill
\centering{
\hspace{-2.5mm}\resizebox{83mm}{!}{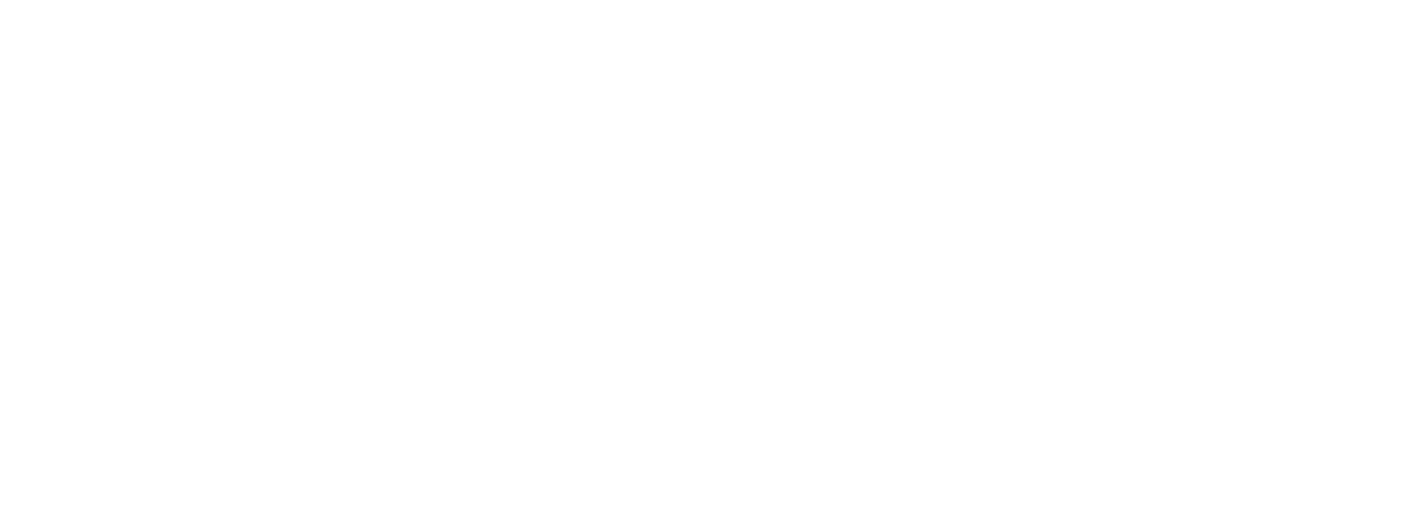}
}
\caption{In the path inference problem, we aim to predict a path suffix $s = (v_{t+1}, \ldots, v_{t+h})$ given a graph and a prefix $p = (v_1, \ldots, v_t)$ (left). In the generalized formulation, instead of a prefix we are given a trajectory $\phi$ encoding the approximate position of a subset of nodes in the prefix (right).}
\label{fig:path_trajectory}
\end{figure*}

From a deep learning perspective, the main challenge we face is reconciling graph-based approaches with sequential data.
Graph convolutional networks (GCN) have exhibited a measure of success at predicting properties of nodes (e.g., the category of a Wikipedia article or the average traffic flow in a road network)~\cite{Bruna,ChebyNet,kipf_welling} or of the graph as a whole (e.g., the solubility of a molecule or the functional similarity of two proteins)~\cite{hamilton2017inductive,xu2018powerful}. The isotropic nature of graph convolution however renders it a poor fit for sequential data.
On the other hand, sequence prediction problems are typically solved with Recurrent Neural Networks (RNN)~\cite{lstm}. These are ideal for pure sequences, such as sentences or timeseries, but do not take into account the graph structure.

\paragraph{Contributions.} This paper proposes \alg, a graph neural network that acts as a generative model for paths. We teach our network to modify a graph so that it encodes the directionality of an observed path prefix. Candidate suffixes are then generated by a non-backtracking walk on the modified graph. \alg's simple form comes with a number of benefits. Inference can be done efficiently and in closed-form. In addition, the network can be trained to estimate the true path likelihood from very little data. 

We evaluate the validity of our approach in two diverse tasks: extrapolating GPS traces on a road network~\cite{traj-rnn} and predicting the Wikipedia article a player is navigating towards in the game of Wikispeedia~\cite{wikispeedia}.
As confirmed by our experiments, \alg is highly anisotropic in its operation, despite being based on graph convolution. It also compares favorably to state-of-the-art RNN and other baselines that do not fully exploit the graph structure. A case in point, \alg identifies the correct path $\sim$28\% more frequently than the best RNN in the GPS trace experiment and achieves an 8-fold target accuracy improvement in the Wikispeedia dataset for a horizon of three hops.

\section{The path inference problem}
\label{sec:problem}

Suppose that there exists an agent\footnote{Though in some cases an actual agent might not exist, the path inference problem becomes more intuitive if we pretend that it does.} that navigates a directed graph $G = (\cV, \cE)$ consisting of $n = |\cV|$ nodes and $m = |\cE|$ edges. The agent occupies a single vertex at a time $t$ and it may take one step to move between node $v_i$ to $v_j$ whenever a directed edge $e_{i \to j} \in \cE$ exists. 
Its position across time is summarized by the traversed {path} $(v_{1}, v_{2}, \ldots, v_{t})$, which is a sequence of nodes that are pairwise adjacent
(see Figure~\ref{fig:path_trajectory}, left). 
We write $\fedge{t}$ and $\bedge{t}$ to refer respectively to the edges in the forward and backward path order: forward edge  $\fedge{t}$ goes from $v_{t} \to v_{t+1}$, whereas $\bedge{t}$ traverses the path in the opposite direction, from $v_{t} \to v_{t-1}$.

\paragraph{Path inference.} The problem we consider entails estimating the likelihood $\prob{s\,| \, p,G}$ of a \textit{suffix} path $s = (v_{t+1}, \ldots$, $v_{t+h})$ given a \textit{prefix} path $p = (v_{1}, \ldots, v_{t})$ on graph $G$.
The likelihood may additionally depend on an assortment of available features relating to nodes, edges, and the agent itself.
Further, since the number of possible paths increases exponentially with the prediction horizon $h$, it is also important to find efficient ways of (\textit{i}) sampling paths, and (\textit{ii}) identifying the one that has the maximum likelihood. In certain situations, one may also be interested in (\textit{iii}) the marginal likelihood of $v_{t + h}$, i.e., the probability that the agent reaches $v_{t + h}$ after $h$ steps.

\paragraph{Generalization.} The above formulation assumes that the path prefix is exactly known---a requirement that may not be met in practice. To this end, we generalize the path inference problem in two ways.
First, we suppose that we possess only an approximate idea of the agent's position. For every $t$, we represent our knowledge by a vector $\xx_{t} \in \mathbb{R}_{\geq 0}^{n}$, conveniently normalized to have measure one. The $i$-th entry of $\xx_{t}$ is then interpreted as the likelihood that the agent resides at node $v_i$ at step $t$. This comes handy also if we try to extrapolate a path recursively, as subsequent calls take into account the uncertainty of previous decisions.  
Second, we posit that only a subset of the agent's path, called a trajectory, can be observed. Let $\idx$ be a sub-sequence of $(1, 2, \ldots, t)$. A \emph{trajectory} 
$$
\bphi \defeq (\xx_{\tau} : \tau \in \idx)
$$
is then a sub-sequence of $(\xx_{1}, \xx_{2}, \ldots, \xx_{t})$. Since $\bphi$ is always defined in terms of $\idx$, whenever a function has access to $\bphi$ in the following we assume that it also knows $\idx$ (though this remains implicit in the notation).      
With this in place, the \emph{generalized path inference} problem amounts to estimating the likelihood $\prob{s \, | \, \bphi, G}$ of a path suffix $s = (v_{t+1}, v_{t+2}, \ldots, v_{t+h})$ given a trajectory $\bphi$ and the graph $G$.

\section{Finding paths with \alg}
\label{sec:solution}

We wish to construct a generative model for paths. Given an input trajectory and a horizon the model should be able to generate candidate suffixes and inform us of their likelihood.

A key challenge in this pursuit lies in capturing directionality. 
Setting aside the special case of product graphs\footnote{A product graph can be expressed as the graph product of $k$ simpler graphs. For such graphs, one may define $k$ consistent notions of direction, each corresponding to one constituent graph. Examples include the grid and hyper-cube (Cartesian product),  Rook's graph (Tensor product), and King's graph (Strong product).}, most graphs are not imbued with a natural notion of direction. 
This is also the reason why graph convolution is an isotropic operation: from a graph-theoretic perspective, there is no consistent way of ordering or grouping the neighbors of any given node.
Fortunately, what we refer to in jest as ``curse of directionality'' can be broken 
if one combines the graph structure with additional information, such as a path prefix:
given a path of length $t$, every node becomes capable of separating its neighbors in up to\footnote{The exact number depends on the level-sets of the distance function for every node in the path.} $3 t$ groups depending on whether they are closer, equidistant, or further from each node in the path.

Armed with this intuition, our approach will be to train a graph neural network called \alg to {encode} all available information about a path prefix into a \emph{latent graph} 
$$ \Phi \defeq (\cV, \cE, w_\bphi), ~~ \text{with} ~~ w_\bphi = f_{\theta}(G, \bphi). $$ 
Though $\Phi$ shares the same vertex and edge sets as $G$, its edges are re-weighted so as to point towards the directions the agent is most likely to follow. 
We will then approximate the likelihood of any suffix $s$ by the graph-dependent model 
$$ \prob{ s \, | \, h, \bphi, G } \simeq g(s, \xx_{t}, \Phi), $$
where $\xx_{t}$ captures the last known position of an agent. The notation above indicates that the model sees $\Phi$ instead of $G$.

\subsection{Capturing direction} 
We train a graph neural network $f_{\theta}(G, \bphi)$ to predict the most likely direction in the vicinity of every node. The network assigns a weight to every directed edge by combining available features with a system of learned pseudo-coordinates (equivalently embeddings), capturing the relation between every node and observation.

The pseudo-coordinate vectors $\cc_i = [\cc_{i,1}, \ldots, \cc_{i,|\bphi|}]$ characterizing each node $v_i$ are jointly learned by a graph convolutional network (GCN) of $k$-layers: 
$$
    \cc_{i,\tau} = [\text{GCN}(\xx_{\tau})]_{i,:} ~~ \text{for} ~~ \tau \in \idx.
$$ 
The GCN we employ computes the $\tau$-th pseudo-coordinate vector in parallel for all nodes in a recursive manner:
$$
    [\mX_k]_{i,:} = \text{ReLU}\hspace{-.5mm}\left( \sum_{v_j \in \cV} w(e_{j \to i}) \, [\mX_{k-1}]_{j,:} \mW_k \right),
$$
with $\mX_0 = \xx_{\tau}$.
Each edge weight is initialized by a multi-layer perceptron
$
    w(e_{i \to j}) = \text{MLP}(\ff_i, \ff_j, \ff_{{i \to j}})
$
taking as input the features of nodes $v_i$ and $v_j$ as well as those corresponding to edge $e_{i \to j}$.
Finally, the weight $ w_\bphi(e_{i \to j})$ of every directed edge is decided by a simple network that predicts the most likely direction. We use once more a multi-layer perceptron  
$$ 
    z_{i \to j} = \text{MLP}( \cc_i, \cc_j, \ff_i, \ff_j, \ff_{i \to j} ),
$$
followed by a soft-max over outgoing edges
$$
    w_\bphi(e_{i \to j}) = \dfrac{ z_{i \to j}}{ \sum_{ v_l \in \cV} z_{i\to l}}.
$$
The latter ensures that the weights of all out-going edges of each $v_i$ can be interpreted as a categorical distribution. 
Akin to skip connections in residual networks, we reuse features in both MLP so as to facilitate training.

\subsection{A generative model with short-term memory}

We opt for a generative model that performs a \textit{non-backtracking walk} on the latent graph. 
Akin to a random walk, the model assumes that the agent traverses each forward edge $\fedge{\tau}$ from $v_{\tau}$ to $v_{\tau+1}$ with probability proportional to the learned edge weight $w_\bphi( \fedge{\tau})$. The main difference is that the walk cannot return to its previous position (i.e., $\fedge{\tau} \neq \bedge{\tau}$). 

Using a model with short-term memory has two interesting consequences. First, the graph neural network is encouraged to find a meaningful latent graph, capturing the directionality of the path. At the same time, inference can be done in closed-form (and often efficiently), greatly simplifying training. We provide three examples in the following:

\paragraph{Suffix likelihood.} 
The likelihood of a path suffix $s = (v_{t+1}, \ldots, v_{t+h})$ is  
\begin{align}
    g(s, \xx_t, \Phi) 
    &= \sum_{v_{t} \in \cV} \left(\prod_{\tau = t}^{t+h-1} \hspace{-1.0mm} p_{\tau} \right) [\xx_{t}]_{v_{t}},  \notag 
\end{align}
where $\xx_{t}$ captures the last known position of an agent and the non-backtracking probabilities $p_{\tau}$ are 
$$
    p_{\tau} \defeq 
    \begin{dcases} 
    w_\bphi( \fedge{\tau}) &  \text{if } \tau = t, \\
    \dfrac{ w_\bphi( \fedge{\tau}) }{ 1 - w_\bphi(\bedge{\tau})} & \text{if } \fedge{\tau} \neq \bedge{\tau} \text{ and } \tau > t,\\
    0       & \text{o.w. } 
    \end{dcases}
$$
It can be seen that the computational complexity grows linearly with the support of $\xx_t$ and the horizon $h$. Hence, when aiming to extrapolate paths we can efficiently train our network by minimizing directly the negative log-likelihood (NLL) of the true suffix.

\paragraph{Target likelihood.} Alternatively, we can train our network to predict the distribution $\xx_{t+h}$ of the target over a known horizon.
Following \cite{Kempton2016NonBacktrackingRW},
let $\mP_{\bphi}$ be the $m \times m$ non-backtracking matrix with 
$$
    [\mP_{\bphi}]_{e_{i \to j}, e_{k \to l} } = 
    \begin{dcases} 
    0       & \text{if }  j \neq k \text{ or } i = l, \\ 
    \dfrac{ w_\bphi( e_{k \to l} ) }{ 1 - w_\bphi( e_{k \to i} ) } & \text{o.w. } 
    \end{dcases}
$$
Further, define the $m \times n$ matrix $\mB_{\bphi}$, with $[\mB_{\bphi}]_{e_{i \to j}, k} = 0$ if $k \neq i$ and $w_\bphi( e_{k \to j} )$ otherwise.  
The marginal distribution $\hat{\xx}_{t+h}$ of the non-backtracking walk on $\Phi$ after $h$ steps can be written as 
\begin{align}
    \hat{\xx}_{t+h}
    &= \mB_{\bphi}^+ \, {\mP_{\hspace{-0.5mm}\bphi}^{\hspace{0.5mm}h}} \, \mB_{\bphi} \, \xx_{t}, \notag 
\end{align}
where due the special sparsity structure of $\mB_{\bphi}$ (its rows have disjoint support),
the pseudo-inverse $\mB_{\bphi}^+$ is, up to normalization, equal to $\mB_{\bphi}^\top$. The computational complexity is thus linear $O(m h)$ w.r.t. the number of edges and horizon.
The network can be trained by minimizing the cross entropy between $\hat{\xx}_{t+h}$ and $\xx_{t+h}$ or any other measure between distributions. 

\paragraph{Most likely suffix.} The maximum likelihood suffix over a given horizon can be identified by Monte-Carlo sampling or deterministically. 
Suppose that the agent resides at node $v_t$ almost surely. Further, let $H_{\phi}$ be the weighted directed graph whose adjacency matrix is $\log(\mP_{\phi})$ (the logarithm is applied only to non-zero entries of $\mP_{\phi}$). 
Since the nodes of $H_\phi$ correspond to edges in $G$, a suffix $s$ can also be seen as a path $(\fedge{t}, \ldots, \fedge{t+h-1})$ on $H_\phi$. Moreover, as a consequence of the transformation, the weight of $s$ in $H_{\phi}$ is the same as its log-likelihood:
\begin{align}
    \log(g(s, \bm{\delta}_t,\Phi)) 
    &= \log \left( w_{\phi}({\fedge{t}})  
    \prod_{\tau = t}^{t+h-2} [\mP_{\phi}]_{\fedge{\tau}, \fedge{\tau+1}}\right) \notag \\
    &= \log( w_{\phi}(\fedge{t})) + \sum_{\tau = t}^{t+h-2} \log\left( [\mP_{\phi}]_{\fedge{\tau}, \fedge{\tau+1}}\right), \notag
\end{align} 
where $\bm{\delta}_t$ is a dirac centered at $v_t$.
The most likely suffix is therefore identified in a deterministic manner by performing a best-first traversal starting from $v_t$ and searching for the maximum-weight path of length $h$. The computational complexity is $O(\Delta^{h})$, where $\Delta$ is a bound on the maximum degree in the $h$-hop neighborhood of $v_t$ (the bound is  tight for a perfect $\Delta$-ary tree of depth $h$ with all edge weights being equal).

\section{Experiments}
\label{sec:experiments}

Our goal is two-fold. First, in Section 4.1 we wish to confirm that \alg can capture the directionality of (straight) paths in the edges of the latent graph. In addition, we are interested in evaluating the generality of our approach and its performance with real data. This is pursued by taking on two diverse tasks: GPS trace extrapolation in Section~\ref{ssec:gps} and user-navigation on a knowledge network in Section~\ref{ssec:wikispeedia}. Code and datasets are publicly available at \textcolor{blue}{\url{https://github.com/jbcdnr/gretel-path-extrapolation}}.

\subsection{Can \alg learn a straight path?}
\label{ssec:synthetic}

We constructed a toy experiment to qualitatively test whether \alg has the capacity to capture directionalilty in the Euclidean sense.
Specifically, we generated straight trajectories on a random graph built to approximate a plane (by uniformly sampling $n=500$ points in $[0,1]^2$ and applying a $10$-NN construction). The trajectories were obtained by mapping straight lines to the closest nodes and sub-sampling the resulting path. 

\begin{figure}[t!]
\centering
  \begin{subfigure}[b]{0.245\linewidth}
    \includegraphics[width=\textwidth, trim={1.0mm 3mm 5mm 3mm}, clip]{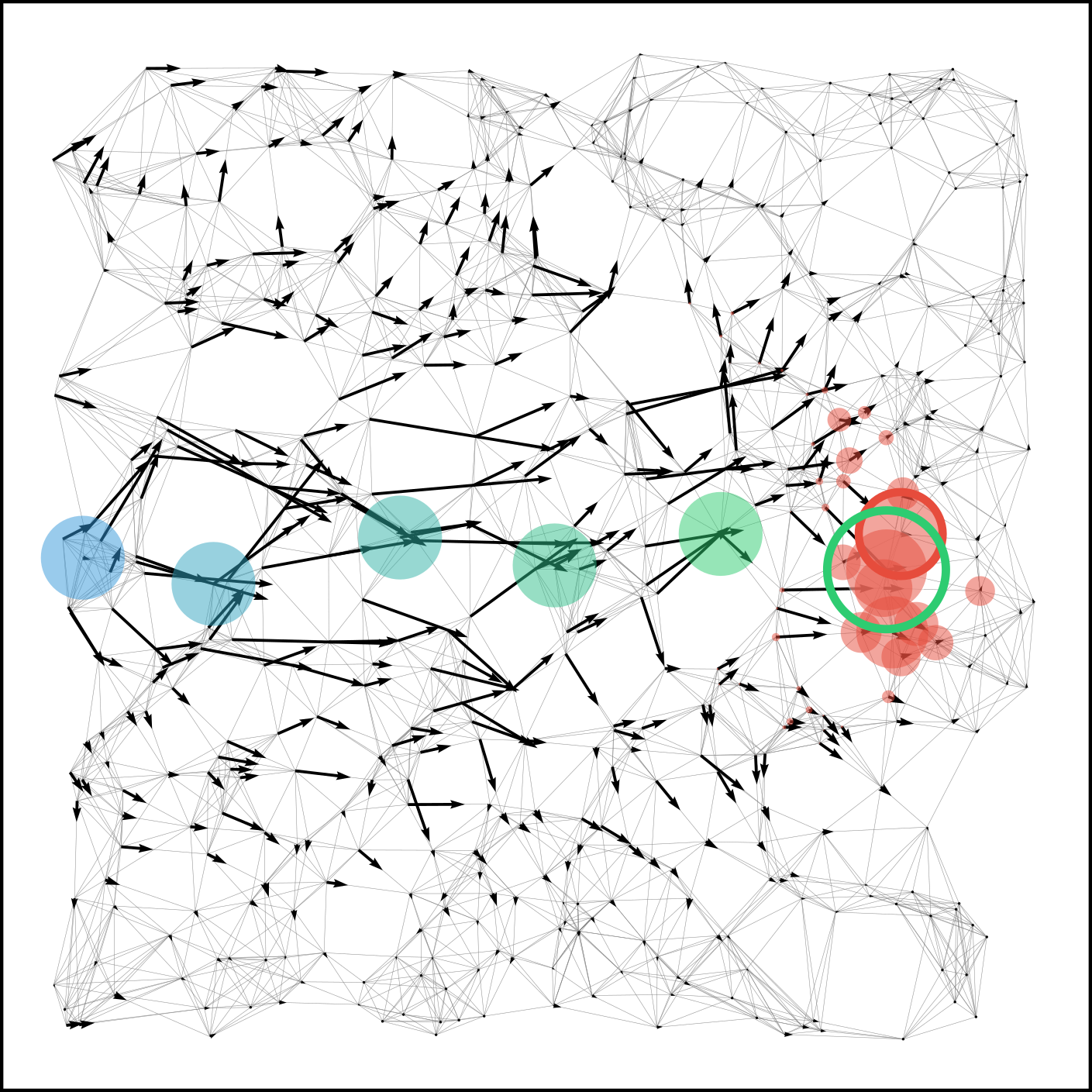}
  \end{subfigure}
  \begin{subfigure}[b]{0.245\linewidth}
    \includegraphics[width=\textwidth, trim={1.0mm 3mm 5mm 3mm}, clip]{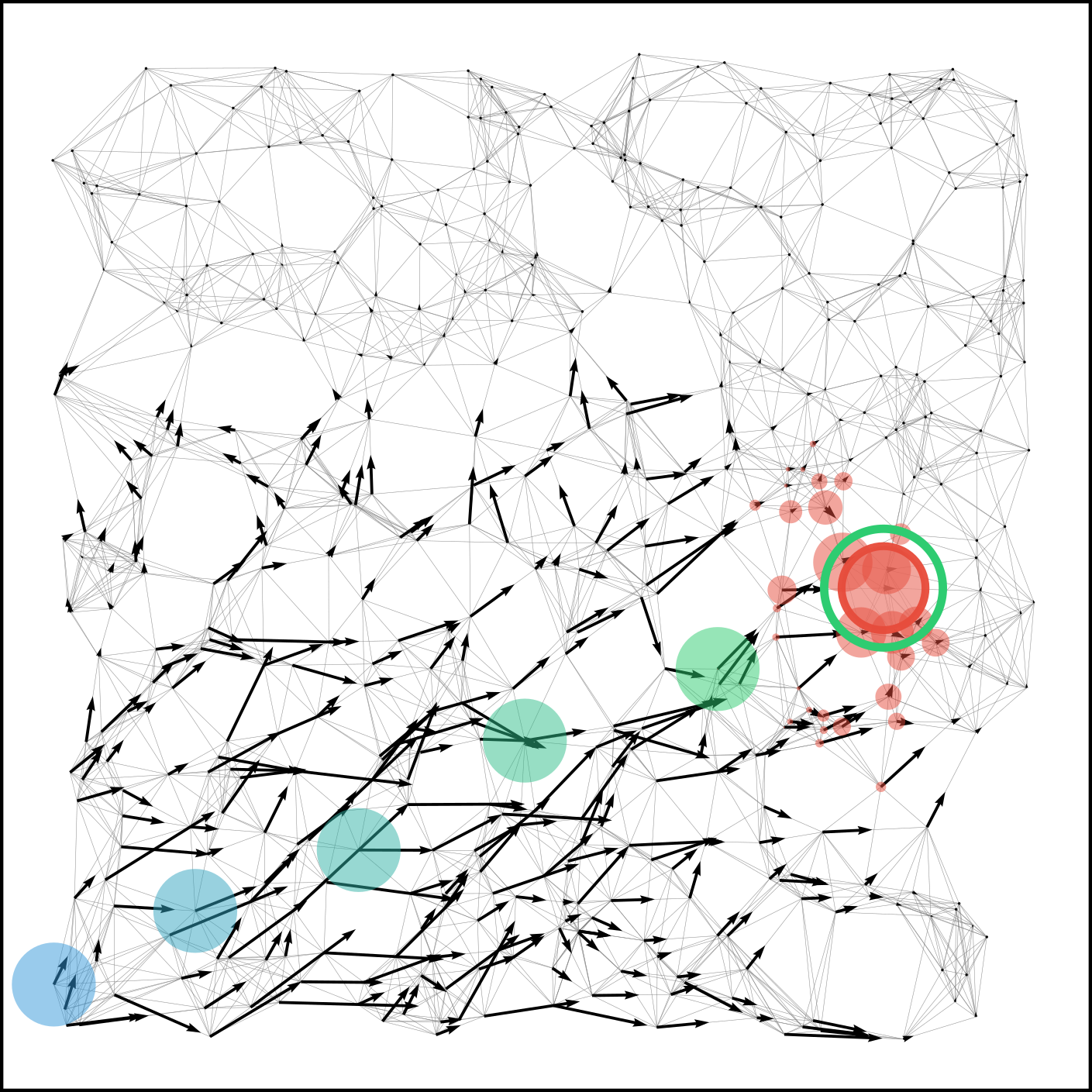}
  \end{subfigure} 
%   \vspace{3mm}
  \hfill
  \begin{subfigure}[b]{0.245\linewidth}
    \includegraphics[width=\textwidth, trim={1.0mm 3mm 5mm 3mm}, clip]{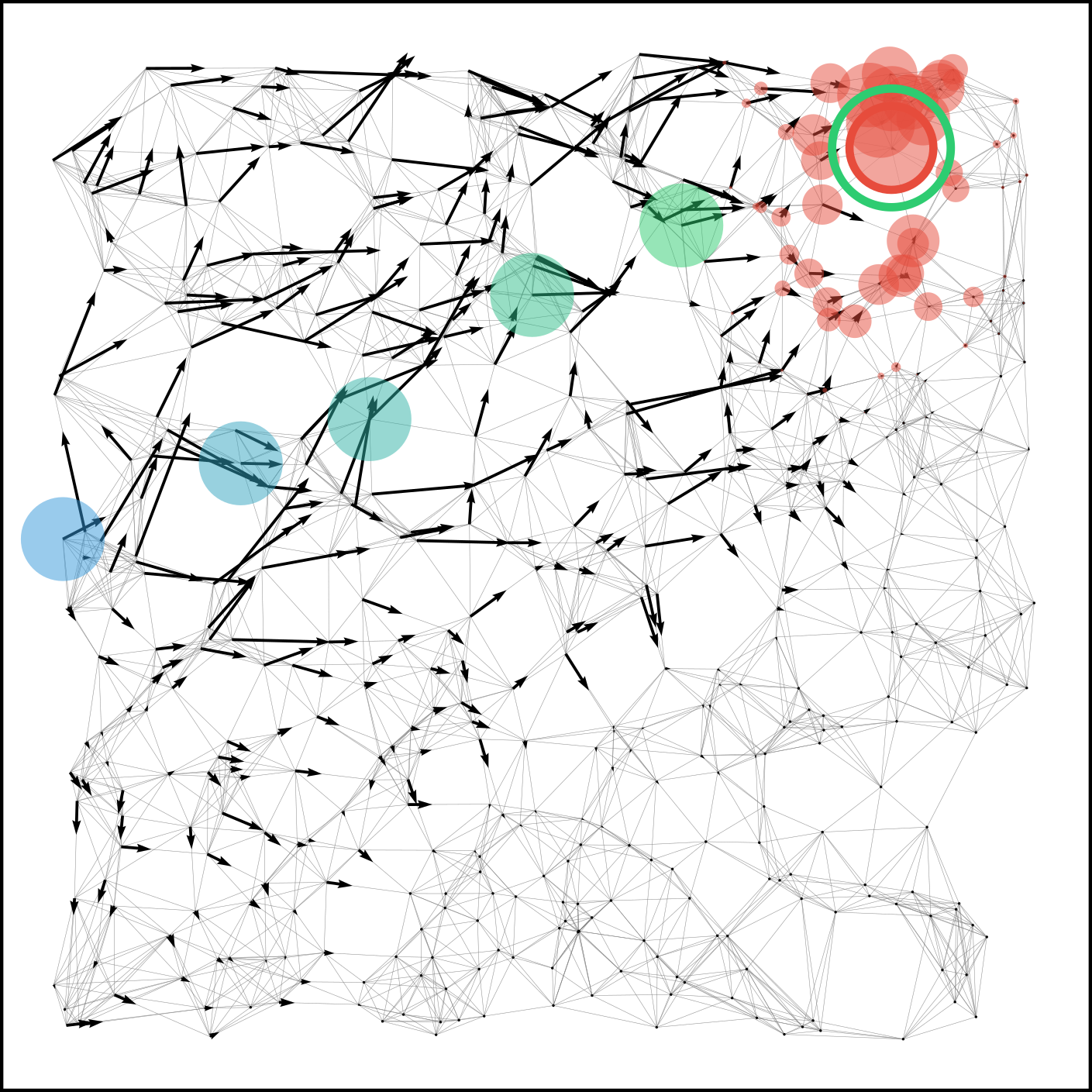}
  \end{subfigure}
  \begin{subfigure}[b]{0.245\linewidth}
    \includegraphics[width=\textwidth, trim={1.0mm 3mm 5mm 3mm}, clip]{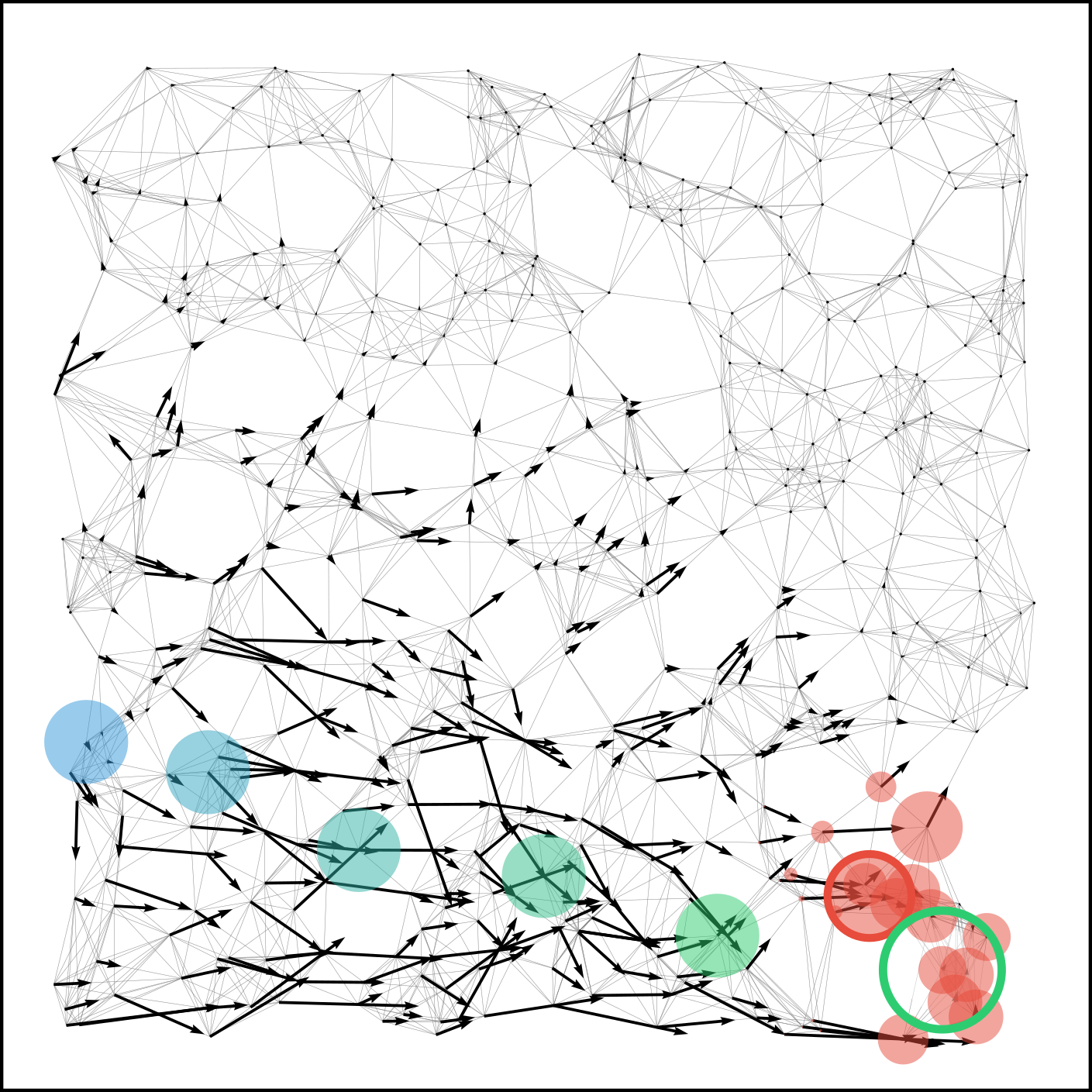}
  \end{subfigure}  
  \caption{Extrapolation of a straight trajectory. The input trajectory is visualized by disks whose color varies from blue to green. True target is highlighted by a green circle. \alg's predicted target distribution is shown with red disks and the maximum likelihood target with a red circle. Arrows indicate chosen edges at each node, length represents confidence.}
  \label{fig:planar}
\end{figure}

Four typical runs of the experiment are shown in \Cref{fig:planar}. Given a trajectory (disks from blue to green), \alg was trained to predict the target (green circle) by minimizing a target cross entropy loss.
The task is non trivial as \alg is not given the positions of the nodes.
In addition, the graph differs from a regular grid and does not offer a good approximation of the underlying Euclidean space.

As seen in Figure~\ref{fig:planar}, most of the probability mass (red disks) of the predicted distribution $\hat{\xx}_{t+h}$ is concentrated close to the target. Moreover, as intended, the direction of the trajectory is encoded into the edge weights of $\Phi$, despite the sampling irregularity (note that a black arrow indicates the most significant edge at each node). 
In the rightmost figure, due to the existence of a hole between the end of the trajectory and the target, the graph neural network assigns small likelihood to the correct target. We hypothesize that the phenomenon is exaggerated by the graph being dense near the north hole boundary, which causes the learned distance metric (implied by the learned pseudo-coordinates) to locally deviate from the Euclidean distance. 
It is also intriguing to observe that, whereas in the leftmost figure \alg does not identify correctly the target, the neural network's answer presents a visually plausible alternative.    

% ---------------------------------------------------------------------------
\subsection{GPS trace extrapolation}
\label{ssec:gps}

In the GPS trace extrapolation problem, we observe a prefix of ordered GPS locations emitted by a driver moving on the road network. 
Two distinct objectives can be addressed: (\textit{i})~predict the position of the driver in $h$ seconds, or (\textit{ii}) predict the following roads that the driver will follow.

\begin{figure}[t]
  \begin{subfigure}[b]{0.24\linewidth}
    \includegraphics[width=\textwidth]{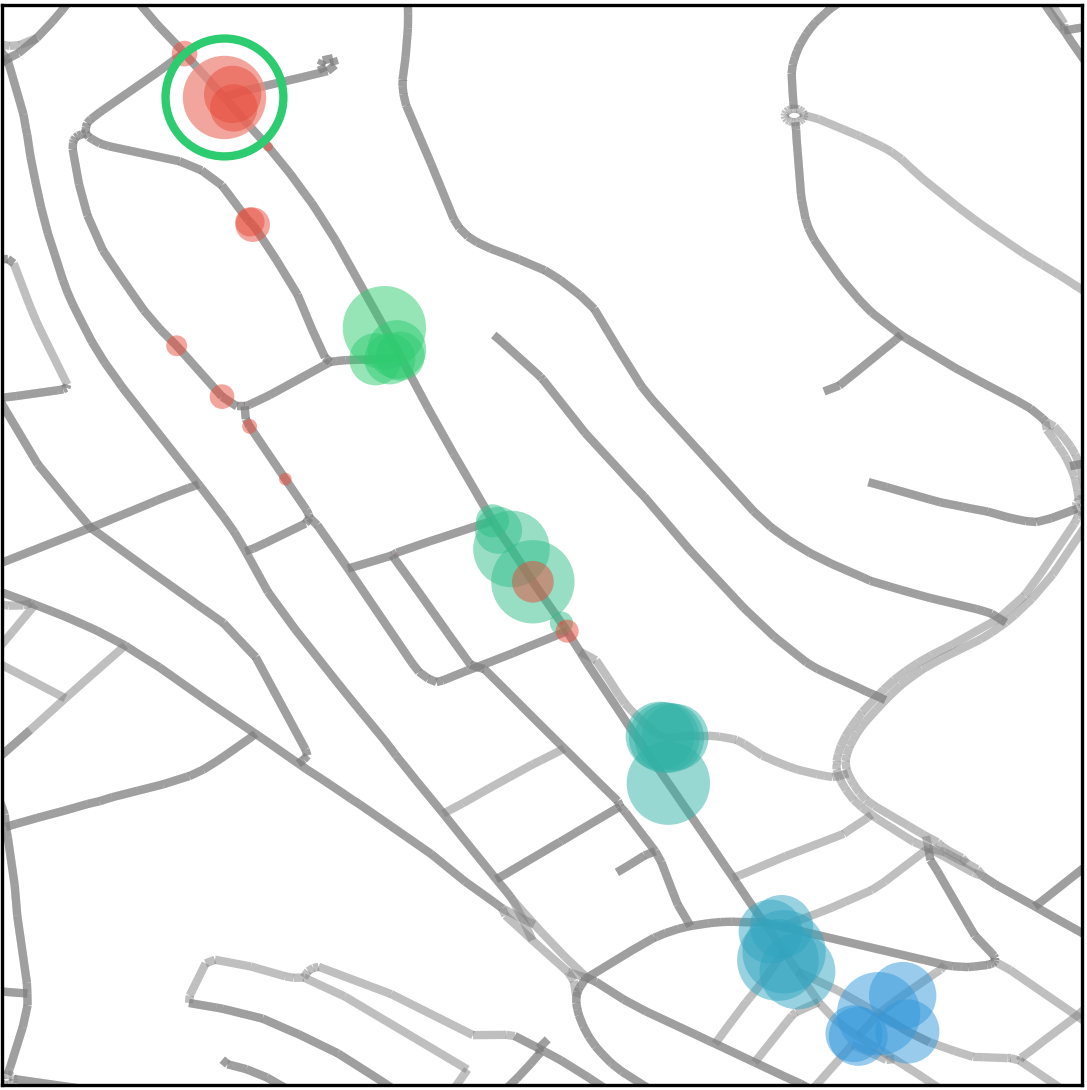}
  \end{subfigure}
  \hfill
  \begin{subfigure}[b]{0.24\linewidth}
    \includegraphics[width=\textwidth]{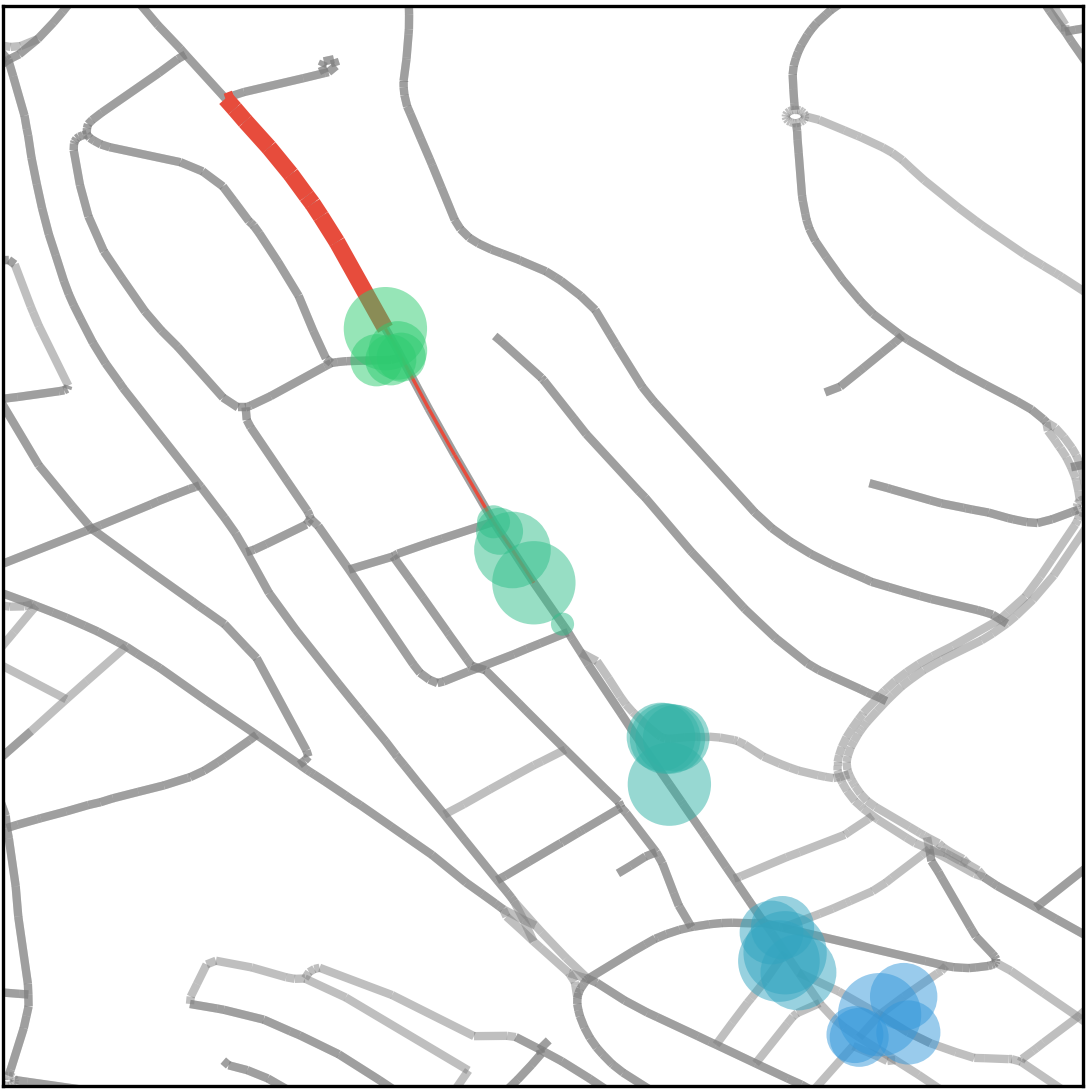}
  \end{subfigure}
 \hfill
  \begin{subfigure}[b]{0.24\linewidth}
    \includegraphics[width=\textwidth]{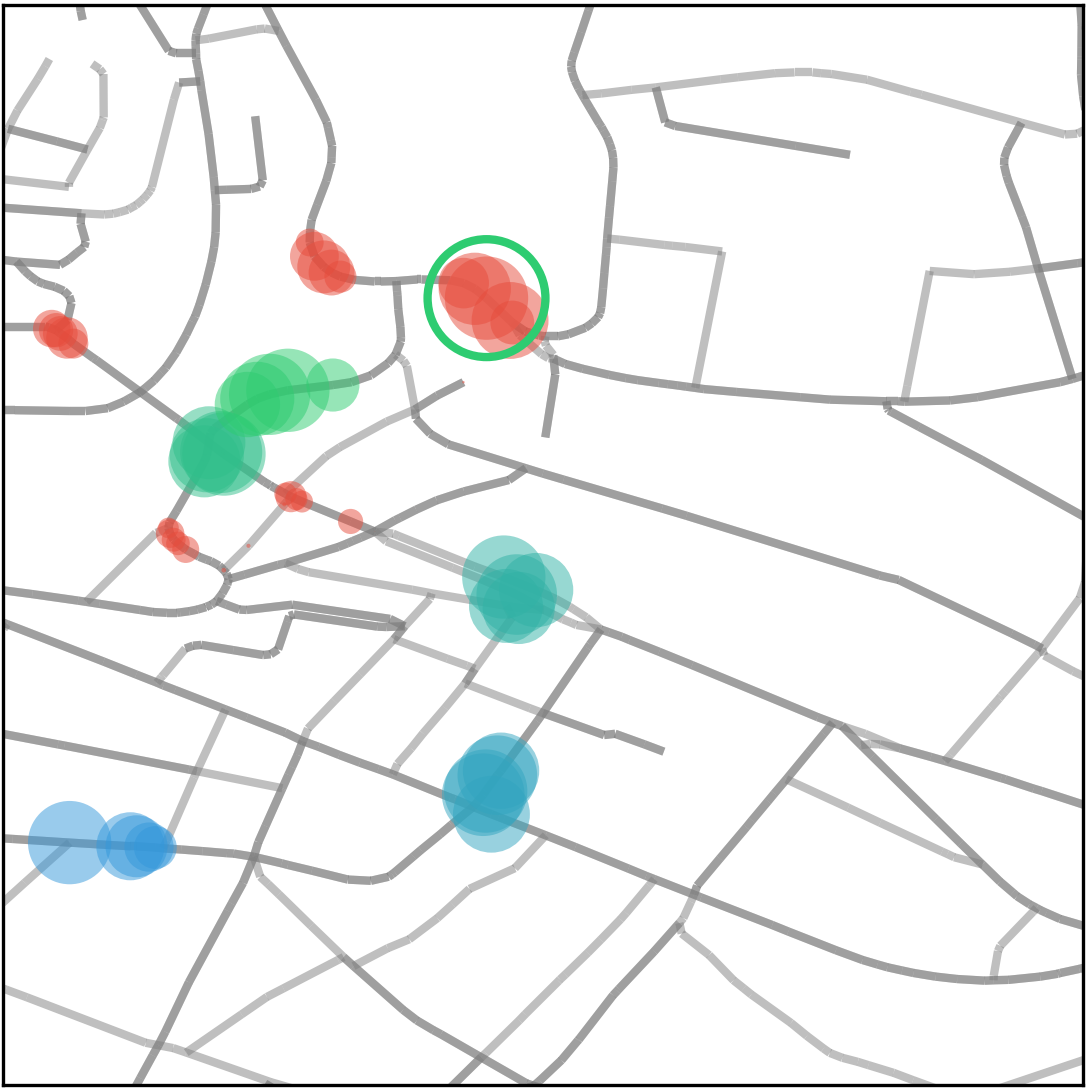}
  \end{subfigure}
  \hfill
  \begin{subfigure}[b]{0.24\linewidth}
    \includegraphics[width=\textwidth]{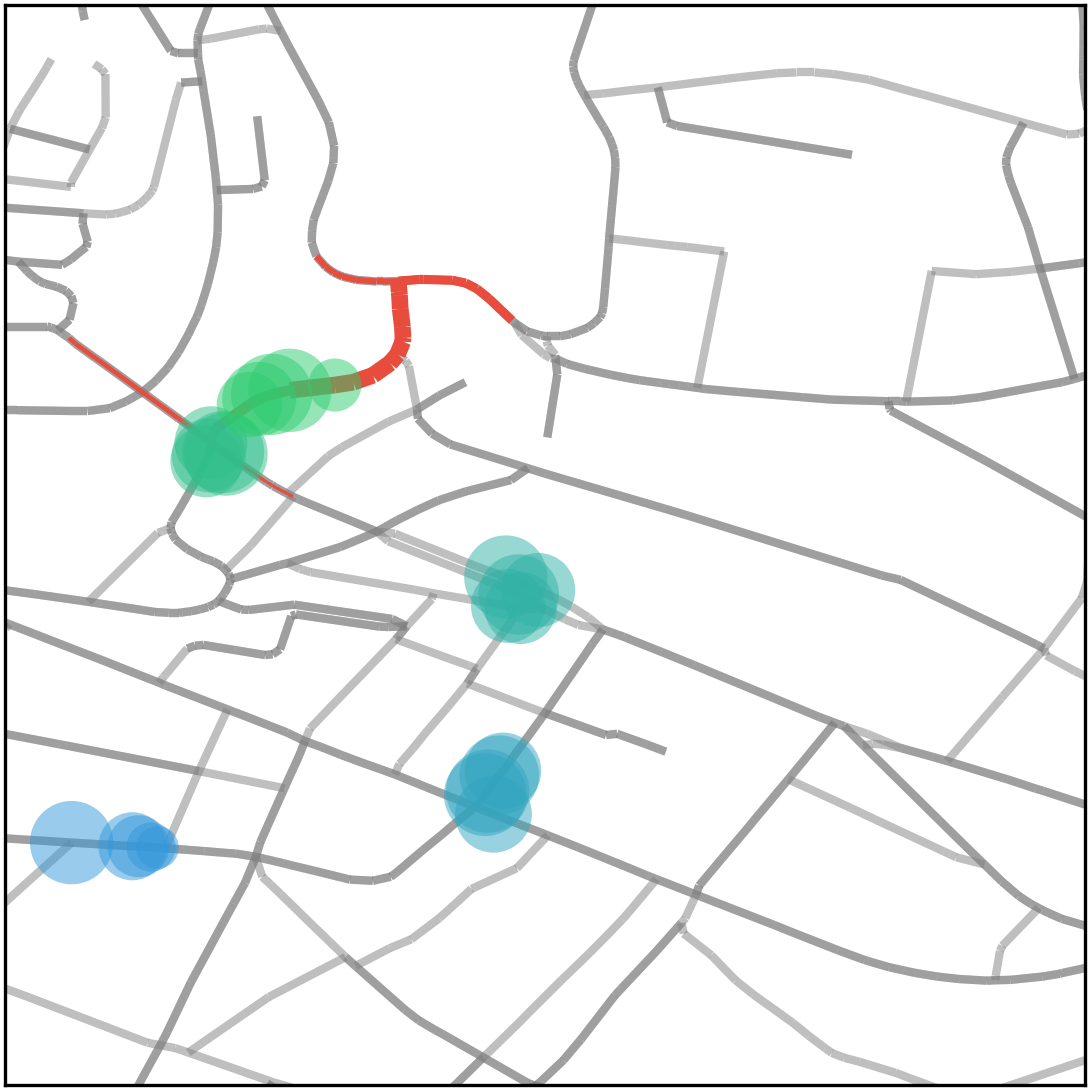}
  \end{subfigure}
  \caption{Two examples of GPS trace extrapolation. Past trajectories are composed of 5 observations (blue to green). \emph{Left:} target distribution as red disks, green circle is the true target. \emph{Right:} sampled future suffixes in red.}
  \label{fig:maps}
\end{figure}

\Cref{fig:maps} illustrates visually the two scenarios. Two trajectories are shown (blue to green filled circles) along with the output of \alg for each objective (red): the target distribution is on the left and the sampled likely suffixes are on the right. Larger markers/bolder lines correspond to more likely targets according to our model. The goal is to predict the true target (highlighted by a green circle) and extrapolate the trajectory towards it.

\paragraph{Existing solutions.} The classical approach to learn patterns from navigation-traces is to model them by a Markov Decision Process and learn the transition probabilities from the observed data.
The Markovian property can also be relaxed by taking into account multiple steps at a time (akin to $n$-grams). The main issue with such approaches is their sample complexity---accurately estimating the probability of rare state transitions requires prohibitive amount of data as the number of parameters grows in the best case (i.e., even when $n=1$) linearly with the number of edges. 
More recently, the GPS extrapolation problem was solved by a recurrent neural network (RNN), achieving state-of-the-art performance~\cite{traj-rnn}. The architecture in question resembles a standard RNN with the main difference that it integrates the choice restrictions induced by the road network at each step.  

\paragraph{Experimental setup.}
We ran an experiment based on a small dataset of food deliveries (229 traces) occurring over the OpenStreetMap\footnote{\url{https://openstreetmap.org}} road network of Lausanne (18156 nodes, 32468 edges). We mapped the GPS coordinates to the $k=5$ closest intersection nodes. The trajectories were preprocessed to have consecutive observations at least 50 meters apart. The min/max/median node degree was 1/6/2.
It is important to note that a GPS trace corresponds to a sequence of noisy physical locations and not a sequence of adjacent nodes.
Nevertheless, all methods discussed so far require a path in order to function properly.
Hence, as an extra pre-processing step, for all baselines the GPS traces were mapped to paths using a Hidden Markov Model (HMM)~\cite{hidden-markov-map-matching-noise-sparseness}. 
To test the versatility of our approach, we did not employ the map-matching algorithm with \alg, but provided it with the raw trajectories as input. 
We used a non-parametric diffusion in the encoder as a learned GCN did not improve the performance.
We trained the RNN based models from \cite{traj-rnn} on one jump extrapolation objective instead of the full suffix to expose the same samples as our method.

\paragraph{Results.} Table~\ref{tab:results_gps} reports extrapolation accuracy w.r.t. three measures. First, \emph{choice accuracy} measures how accurate are the decisions of an algorithm at each crossroad of the ground-truth path connecting $v_t$ and $v_{t+h}$, as extracted by the HMM. We computed the choice accuracy on only nodes whose degree was at least 3, as the decision is trivial otherwise. As seen by the accuracy of a uniform and non-backtracking random-walk, most crossroads encountered had degree 3, leading to a random decision succeeding $\sim$31.5\% of the time. 

\begin{table}[t]
    \centering
% \resizebox{0.5\linewidth}{!}{%
    \begin{tabular}{@{}l c r@{ }l r@{ }l r@{ }l @{}}
        \toprule
        Model&Loss&\multicolumn{2}{c}{Choice accuracy}&\multicolumn{2}{c}{$\prob{v_{t+h}}$}& \multicolumn{2}{c}{NLL} \\
        \midrule
        %
        % BASELINES
        %
        
        Uniform &&31.5&& 0.035 && 2.40&\\
        Uniform NB&&48.8& & 0.133 && 1.89&\\
        \midrule
        \multirow{2}{*}{\alg}
        &target
        &74.2 &$\pm$ 1.4  & \textbf{0.199}& $\pm$ 0.003&\textbf{1.28}& $\pm$ 0.04\\
        &NLL&
        68.8 &$\pm$ 2.2 &  \textbf{0.199} &$\pm$ 0.004 & 1.50& $\pm$ 0.04\\
        \midrule
        %
        % RNN TRAJ REPRODUCE
        %
        CSSRNN* 50&\multirow{4}{*}{NLL}&73.9&$\pm$ 1.6&  && 1.67&$\pm$ 0.08\\
        CSSRNN* 4 &                    &66.2&$\pm$ 2.6&  && 1.57&$\pm$ 0.01\\
        LPIRNN* 50&                    &74.2&$\pm$ 3.1&  && 1.53&$\pm$ 0.06\\
        LPIRNN* 4 &                    &\textbf{75.0}&$\pm$ 2.3&  && 1.87&$\pm$ 0.02\\
        \bottomrule
    \end{tabular}%
% }
    \caption{Results of GPS trace extrapolation on test dataset. Choice accuracy (\%) is computed at non trivial intersections only (more that two outgoing roads). Target probability is not given for the RNN. We use an asterisk to indicate which algorithms have access to the road coordinates.}
    \label{tab:results_gps}
\end{table}

RNN models~\cite{traj-rnn} reached a good choice accuracy on the test set, slightly outperforming \alg (the difference is smaller than the standard-deviation across 5 independent runs). However, they were less competitive in recovering the actual suffix, as measured by the \textit{negative log-likelihood} (NLL) measure. We note that choice accuracy is more lenient with sporadic mistakes as compared to the suffix NLL: the likelihood of a path depends heavily on the worst decision made, whereas this is not true for the former. 
To confirm that the RNN were not affected by overfitting, we repeated the experiment with a smaller hidden size representation of 4 (instead of 50 as proposed by the authors). This brought about only a small improvement to the CSSRNN architecture with the NLL dropping from 1.67 to 1.57 and did not help the LPIRNN.      

Despite achieving moderate choice accuracy, \alg was able to guess the correct suffix $\sim$28\% more frequently than the best RNN: in terms of geometric mean, the two algorithms assigned $0.278$ and $0.216$ probability to the correct suffix, respectively. This is surprising as the RNNs were given a competitive advantage by knowing the road coordinates, whereas \alg did not. Interestingly, the best result was attained when the graph neural network was trained to locate the target, %(maximizing $\hat \xx_{t+h}^\top \xx_{t+h}$), 
even when measuring NLL on the test set. Our hypothesis is that choosing a target loss encourages the neural network to explore alternative suffixes towards the target early on, thus improving training.   

Finally, we report the \textit{target probability} measure $\prob{v_{t+h}}$ corresponding to the average chance an algorithm has to find a node $v_i$ with non-zero $[\xx_{t+h}]_i$ (due to the GPS-to-node mapping procedure, there were five such nodes for each trajectory). We were unable to incorporate this metric for the RNN, as the implementation provided by the authors does not support auto-regressive sampling. In our test set, \alg was able to find the target $\sim$20\% of the time, outperforming simple baselines.

\subsection{User navigation in Wikipedia}
\label{ssec:wikispeedia}

In the Wikispeedia game~\cite{wikispeedia}, human players are called to find a path from a source to a target article by following a sequence of hyperlinks.
Since players can only view the available links locally and guess links on other pages based on their prior knowledge, most paths taken by players differ qualitatively from shortest-paths on the graph of articles~\cite{humain_wayfinding_west}.
Therefore, it is intriguing to determine whether an algorithm can learn to mimic the human routing logic. \textit{In particular, given a path prefix can we predict towards which article the player is navigating towards---or perhaps human choices are too unpredictable?}

Motivated by this question, we trained \alg to predict the target of a navigation prefix among Wikipedia articles. We optimized both objectives, the target probability cross entropy and the suffix negative log-likelihood. Node features were the node in/out degrees (to capture the notion of hubs), while edge feature vectors contained the TF-IDF similarity between source and destination articles of each hyperlink along with the number of times this link was clicked in the training dataset of paths. 
We compare to previous work and baselines which only use article features (e.g. TF-IDF vectors) and local edge features (e.g. node degrees and properties of edges in the prefix).
On the contrary, our method sees the entire graph (not only the local connections of nodes in the suffix), which we will argue is essential to solving this problem.

\paragraph{Previous work.} West and Leskovec~\cite{humain_wayfinding_west} consider two variants of the target prediction task problem: 
(\textit{i}) given a path prefix, the target and a negative target sampled randomly, predict which one is the true target; and 
(\textit{ii}) given a prefix, rank all the possible targets. The code is not publicly available, but we report the accuracy of their 2-targets classifier. 
They extracted carefully tuned features to mimic user way finding, considering for example node degrees for hubs and semantic similarity improvement over the path. Their precision metric to evaluate the ranking model considered sibling articles (same sub-category as the target) as correct predictions, whereas we were less lenient in our evaluation and only considered the prediction correct if the true target was found.

\emph{Other baselines.} We trained a simple predictor based on FastText pre-trained word embeddings of dimension 300~\cite{mikolov2018advances}. 
Article feature vectors $\ff_i$'s were the average of their word representations. Given a prefix $(v_1, \dots, v_4)$, the model computed for each target
$\hat y_j = \sum_{i=1}^4  \ff_i^\top \mW_i \ff_j\,,$
followed by a soft-max to represent a categorical probability over the nodes.
The parameters (i.e., $\mW_i$) were trained using Adam, with the learning rate set to 0.01. This baseline shows what can be achieved without any knowledge of the graph and using only article semantic similarities. For instance, according to FastText, the closest articles to ``Moon'' are ``Mercury'', ``Venus'',  ``Earth's atmosphere'', ``Shackleton crater'' and ``Mars''.

We also compare to two non-parametric versions of \alg{}: (Uniform NB) a non backtracking random walk  run for the distance of the path on the random walk graph starting from $v_t$ and (Reweighted) a walk that has been positively biased towards following frequent links, i.e., those that were commonly favored by players.

\begin{table}[t]
    \centering
% \resizebox{0.5\linewidth}{!}{%
    \begin{tabular}{@{}l r@{ }l r@{ }l r@{ }l c r@{ }l r@{ }l r@{ }l r@{ }l@{}}
        \toprule
        &\multicolumn{6}{c}{precision@1}&~&\multicolumn{6}{c}{2-targets accuracy}\\
        \cmidrule{2-7} \cmidrule{9-14}
        Path length $n$&\multicolumn{2}{c}{5}&\multicolumn{2}{c}{6}&\multicolumn{2}{c}{7}&&\multicolumn{2}{c}{5}&\multicolumn{2}{c}{6}&\multicolumn{2}{c}{7}\\
        \midrule
        Uniform NB&
        1.9&&0.1&&0.0&&&
        49.6&&67.3&&58.2&
        \\
        Reweighted&
         15.9 && 3.8 && 0.6 &&& 
         77.1 && 84.2 && 79.8&
        \\
        FastText&
        3.0 &$\pm$ 0.1&0.7&&0.2&&&
        68.9 &$\pm$ 0.5&70.8 &$\pm$ 0.5&68.2 &$\pm$ 0.6
        \\
        West, 2012&
         && && &&&
        80\phantom{.1}&&84\phantom{.1}&&80\phantom{.1}&
        \\
        \alg target &        
        \textbf{19.5} &$\pm$ 1.3&\textbf{6.2} &$\pm$ 0.3&\textbf{4.9} &$\pm$ 0.3&&
        \textbf{82.2} &$\pm$ 0.3&\textbf{89.2} &$\pm$ 0.3&\textbf{84.6} &$\pm$ 0.3
        \\
        \alg NLL &
        19.0 &$\pm$ 1.4&6.1 &$\pm$ 0.4&4.0 &$\pm$ 1.1&&
        81.6 &$\pm$ 0.3& 88.5 &$\pm$ 0.2& 83.9 &$\pm$ 2.9
        \\
        \bottomrule
    \end{tabular}%
% }
    \caption{Wikispeedia target prediction accuracy (\%) given path prefix of 4 articles. Precision@1 is the ratio of true targets that are assigned the highest probability. 2-targets accuracy is computed on the classification between the true target and a random article at the same distance. The length of the suffix is $n-4$. The symbol $\pm$ indicates one standard-deviation and has been computed on the basis of 5 independent runs. }
    \label{tab:results_wikipedia}
\end{table}

\paragraph{Results.} Table~\ref{tab:results_wikipedia} reports two metrics: \textit{precision@1} measures how often a classifier recovers the actual target, whereas \textit{2-targets accuracy} tests if a classifier can distinguish the true target from a random article selected from those in the same shortest path distance as the true target and given at least once as target (mimicking~\cite{humain_wayfinding_west}).

As seen, \alg achieves between 4\%-and-6\% absolute accuracy improvement over state-of-the-art. We attribute this improvement to it considering each path prefix in light of the full graph between articles: nodes close to the prefix path can be discarded as possible targets as the player would probably have found them or stayed in their close neighborhood. This notion of proximity is not accessible to other methods, while we believe it is crucial in attaining good accuracy. 
The FastText method does incorporate some knowledge of the world not accessible to other methods. However, our experiment suggests that the intrinsic meaning of articles does not suffice to make a good prediction. On the other hand, re-weighting the edges based on how frequently they have been used in the training set is a very effective strategy when trying to predict the next article, but suffers for larger horizons.
A case in point, if \alg was used to suggest to users the next article to look at, it would match their choice 1/5.26 of the time for one hop prediction and 1/20.4 of the time for three hops, whereas the Reweighted baseline would be correct 1/6.28 and 1/166.6 of the time, respectively: for a horizon of three hops, therefore, our method improves target accuracy by $8.16\times$.  

To get a feeling of some of the paths predicted by \alg, we report five hand-picked examples from our test set in Table~\ref{tab:examples}. The first three of these paths are also visualized in Appendix~\ref{appendix}.

\begin{table}[h]
\centering
% \resizebox{\linewidth}{!}{%
\begin{tabular}{ll}
\toprule
 {Prefix}, \textbf{true suffix}, \textit{sampled suffices from \alg}                                                              & $\prob{s\,| \, p,G}$   \\
\midrule
 Lunar eclipse, Sunlight, Electromagnetic radiation, Atom    &          \\
 \textbf{Nuclear fission, Nuclear power}                              &          \\
 \textit{Chemical element, Periodic table}                            & 0.13     \\
 \textit{Nuclear fission, Nuclear power}                              & 0.10     \\
 \textit{Nuclear fission, Nuclear weapon}                             & 0.07     \\
                                                             \midrule
 Latvia, Russia, People's Republic of China, Nepal         &          \\
 \textbf{Himalayas, Edmund Hillary}                                   &          \\
 \textit{Mount Everest, Tenzing Norgay}                               & 0.20     \\
 \textit{Mount Everest, Edmund Hillary}                               & 0.12     \\
 \textit{Himalayas, Yeti}                                             & 0.06     \\
                                                             \midrule
 John Adams, United States, Amtrak, Canadian Pacific Railway &          \\
 \textbf{Rail transport, Train}                                       &          \\
 
 \textit{Rail transport, Train}                                  & 0.15     \\
 \textit{The Canadian, Toronto}                                          & 0.11     \\
 \textit{The Canadian, Rocky Mountains}                               & 0.09     \\
                                                             \midrule
 Marjoram, Juniper berry, Beer, Ethanol &          \\
 \textbf{Distilled beverage, Absinthe}                                       &          \\
 
 \textit{Distilled beverage, Absinthe}                                  & 0.13     \\
 \textit{Alcohol, Phosphorus tribromide}                                          & 0.07     \\
 \textit{Alcohol, Chemistry}                               & 0.06     \\
                                                             \midrule
 Lawrencium, Russia, United States, Publishing &          \\
 \textbf{Newspaper, The Wall Street Journal}                                       &          \\
 
 \textit{Book, Library}                                  & 0.2     \\
 \textit{Book, Novel}                                          & 0.11     \\
 \textit{Newspaper, The Wall Street Journal}                               & 0.07     \\
\bottomrule
\end{tabular}
% }
    \caption{Handpicked examples of the top-3 most likely suffixes according to \alg in the Wikispeedia test set.}
    \label{tab:examples}
\end{table}

\section{Related work}
\label{sec:related_work}

To the extend of our knowledge, this is the first time the generalized path inference problem has been considered.
An interesting relevant work proposed to classify nodes belonging to the shortest path between a source and a target~\cite{deepmind_graph_nets}, but this is a combinatorial problem optimizing a well known graph metric, rather than naturally occurring agents' paths. 

Our refinement of the graph into a latent graph is inspired from their message passing framework. Other specialized graph convolutional network layers, such as Graph Attention Network \cite{graph_attention_networks}, could be also used to tune the edge weights and allow for anisotropic filtering. 
The main difference from these approaches is that we use a non-backtracking walk as a generative model in order to extrapolate paths.

Random walks on graphs have been used previously in a deep learning context in order to sample paths from graphs and extract node representations~\cite{node2vec,deepwalk} using \cite{word2vec}. We can see the pseudo-coordinates as node representations with regard to the observations, but the similarity stops there.

\section{Conclusion}
\label{sec:conclusion}

This paper defines the path inference problem and its generalization to trajectories. 
We proposed a novel graph neural network architecture combining a GCN and a non-backtracking walk generator. Our model refines a graph to capture directionality by conditioning it on a path prefix. The simplicity of the latent representation allows us to sample suffixes efficiently and compute path and target likelihoods.

The path inference problem has remained relatively unexplored, yet it has many applications among which are GPS trace extrapolation and user navigation in information networks, as shown in this work.
We believe that graph neural networks present a promising solution. We are very interested in determining the limits of their ability.

\paragraph{Acknowledgements.} This work was kindly supported by the Swiss National Science Foundation (SNSF) in the context of the project ``\emph{Deep Learning for Graph-Structured Data}'', grant number PZ00P2 179981.

{
\bibliographystyle{unsrt}
\bibliography{bibliography}

\begin{thebibliography}{10}

\bibitem{wikispeedia}
Robert West, Joelle Pineau, and Doina Precup.
\newblock Wikispeedia: An online game for inferring semantic distances between
  concepts.
\newblock In {\em {IJCAI} 2009, Proceedings of the 21st International Joint
  Conference on Artificial Intelligence, Pasadena, California, USA, July 11-17,
  2009}, pages 1598--1603, 2009.

\bibitem{Bruna}
Joan Bruna, Wojciech Zaremba, Arthur Szlam, and Yann Lecun.
\newblock Spectral networks and locally connected networks on graphs.
\newblock In {\em International Conference on Learning Representations
  (ICLR2014)}, April 2014.

\bibitem{ChebyNet}
Micha{\"{e}}l Defferrard, Xavier Bresson, and Pierre Vandergheynst.
\newblock Convolutional neural networks on graphs with fast localized spectral
  filtering.
\newblock In {\em {NIPS}}, pages 3837--3845, 2016.

\bibitem{kipf_welling}
Thomas~N. Kipf and Max Welling.
\newblock Semi-supervised classification with graph convolutional networks.
\newblock {\em CoRR}, abs/1609.02907, 2016.

\bibitem{hamilton2017inductive}
Will Hamilton, Zhitao Ying, and Jure Leskovec.
\newblock Inductive representation learning on large graphs.
\newblock In {\em Advances in Neural Information Processing Systems}, pages
  1024--1034, 2017.

\bibitem{xu2018powerful}
Keyulu Xu, Weihua Hu, Jure Leskovec, and Stefanie Jegelka.
\newblock How powerful are graph neural networks?
\newblock {\em arXiv preprint arXiv:1810.00826}, 2018.

\bibitem{lstm}
Sepp Hochreiter and J\"{u}rgen Schmidhuber.
\newblock Long short-term memory.
\newblock {\em Neural Comput.}, 9(8):1735--1780, November 1997.

\bibitem{traj-rnn}
Hao Wu, Ziyang Chen, Weiwei Sun, Baihua Zheng, and Wei Wang.
\newblock Modeling trajectories with recurrent neural networks.
\newblock In {\em Proceedings of the Twenty-Sixth International Joint
  Conference on Artificial Intelligence, {IJCAI-17}}, pages 3083--3090, 2017.

\bibitem{Kempton2016NonBacktrackingRW}
Mark Kempton.
\newblock Non-backtracking random walks and a weighted {Ihara}'s theorem.
\newblock {\em Open Journal of Discrete Mathematics}, 6(04):207, 2016.

\bibitem{hidden-markov-map-matching-noise-sparseness}
Paul Newson and John Krumm.
\newblock Hidden markov map matching through noise and sparseness.
\newblock pages 336--343, November 2009.

\bibitem{humain_wayfinding_west}
Robert West and Jure Leskovec.
\newblock Human wayfinding in information networks.
\newblock In {\em Proceedings of the 21st International Conference on World
  Wide Web}, WWW '12, pages 619--628, New York, NY, USA, 2012. ACM.

\bibitem{mikolov2018advances}
Tomas Mikolov, Edouard Grave, Piotr Bojanowski, Christian Puhrsch, and Armand
  Joulin.
\newblock Advances in pre-training distributed word representations.
\newblock In {\em Proceedings of the International Conference on Language
  Resources and Evaluation (LREC 2018)}, 2018.

\bibitem{deepmind_graph_nets}
Peter~W. Battaglia, Jessica~B. Hamrick, Victor Bapst, Alvaro
  Sanchez{-}Gonzalez, Vin{\'{\i}}cius~Flores Zambaldi, Mateusz Malinowski,
  Andrea Tacchetti, David Raposo, Adam Santoro, Ryan Faulkner, {\c{C}}aglar
  G{\"{u}}l{\c{c}}ehre, Francis Song, Andrew~J. Ballard, Justin Gilmer,
  George~E. Dahl, Ashish Vaswani, Kelsey Allen, Charles Nash, Victoria
  Langston, Chris Dyer, Nicolas Heess, Daan Wierstra, Pushmeet Kohli, Matthew
  Botvinick, Oriol Vinyals, Yujia Li, and Razvan Pascanu.
\newblock Relational inductive biases, deep learning, and graph networks.
\newblock {\em CoRR}, abs/1806.01261, 2018.

\bibitem{graph_attention_networks}
Petar Velickovic, Guillem Cucurull, Arantxa Casanova, Adriana Romero, Pietro
  Li{\`{o}}, and Yoshua Bengio.
\newblock Graph attention networks.
\newblock {\em CoRR}, abs/1710.10903, 2017.

\bibitem{node2vec}
Aditya Grover and Jure Leskovec.
\newblock Node2vec: Scalable feature learning for networks.
\newblock In {\em Proceedings of the 22Nd ACM SIGKDD International Conference
  on Knowledge Discovery and Data Mining}, KDD '16, pages 855--864, New York,
  NY, USA, 2016. ACM.

\bibitem{deepwalk}
Bryan Perozzi, Rami Al-Rfou, and Steven Skiena.
\newblock Deepwalk: Online learning of social representations.
\newblock In {\em Proceedings of the 20th ACM SIGKDD International Conference
  on Knowledge Discovery and Data Mining}, KDD, pages 701--710. ACM, 2014.

\bibitem{word2vec}
Tomas Mikolov, Ilya Sutskever, Kai Chen, Greg~S Corrado, and Jeff Dean.
\newblock Distributed representations of words and phrases and their
  compositionality.
\newblock In C.~J.~C. Burges, L.~Bottou, M.~Welling, Z.~Ghahramani, and K.~Q.
  Weinberger, editors, {\em Advances in Neural Information Processing Systems
  26}, pages 3111--3119. Curran Associates, Inc., 2013.

\bibitem{ICWSM09154}
Mathieu Bastian, Sebastien Heymann, and Mathieu Jacomy.
\newblock Gephi: An open source software for exploring and manipulating
  networks.
\newblock In {\em International AAAI Conference on Weblogs and Social Media},
  2009.

\end{thebibliography}
}

\appendix
\section{Examples from \alg's Wikispeedia predictions}
\label{appendix}

To provide an intuitive sense of the Wikispeedia path extrapolation task and \alg's behavior, we visualized a subset of the Wikipedia graph, along with the ground-truth path (orange), and the top tree predictions of our algorithm (red, purple, blue in decreasing likelihood) for the examples of Table~\ref{tab:examples}. Aiming to improve visibility, we display in each case only a small fraction of the entire graph, selected to contain the one-hop neighbors of the nodes in each true path. The graph layout selected was the Force-Layout 2 implemented in the gephi software~\cite{ICWSM09154}. Small perturbations were introduced to the positions of some nodes so as to minimize label occlusion.

\begin{figure}[h]
    \centering
    \includegraphics[width=0.73\textwidth]{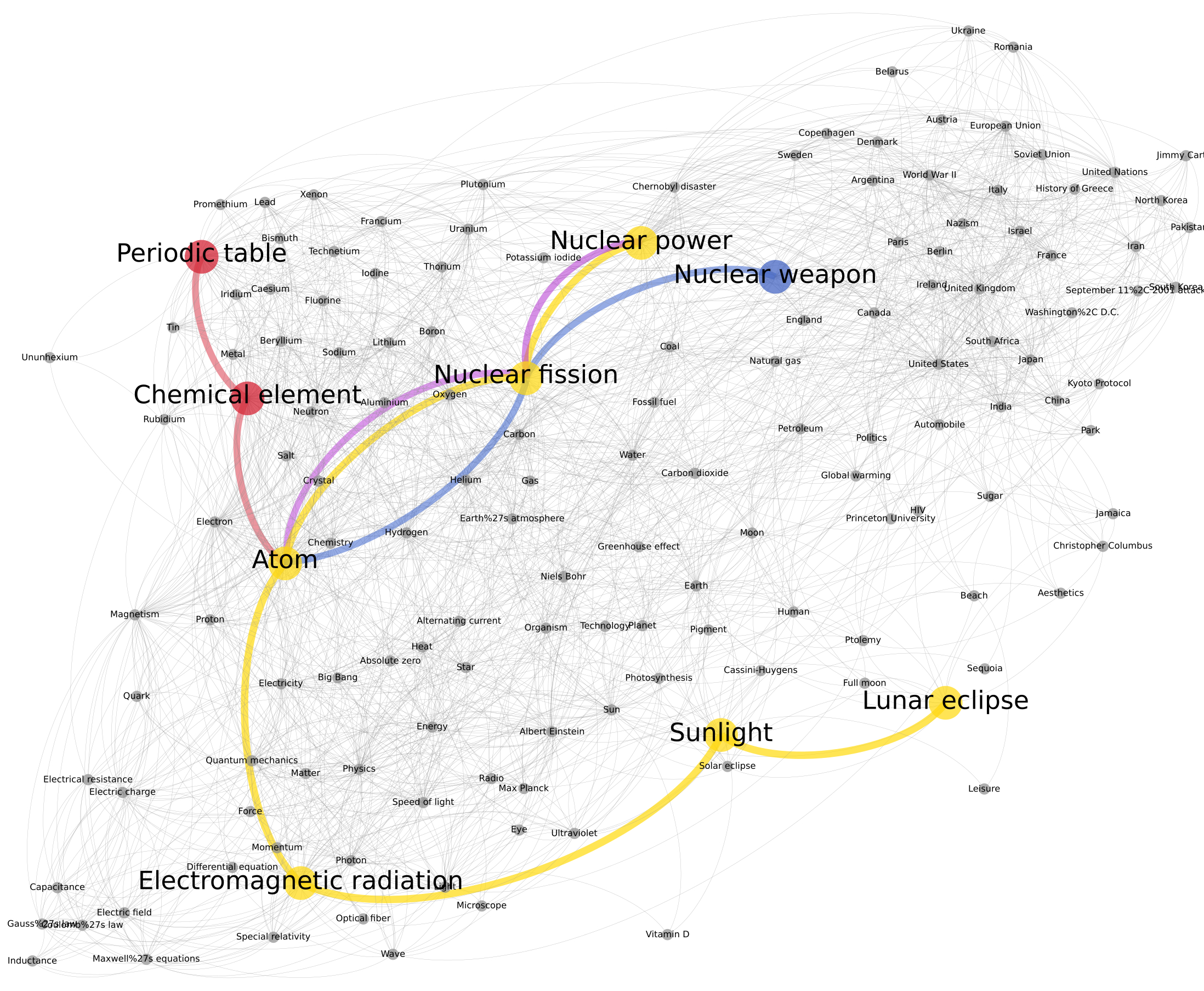}
\end{figure}

\begin{figure}[h]
\centering
    \includegraphics[width=0.86\textwidth]{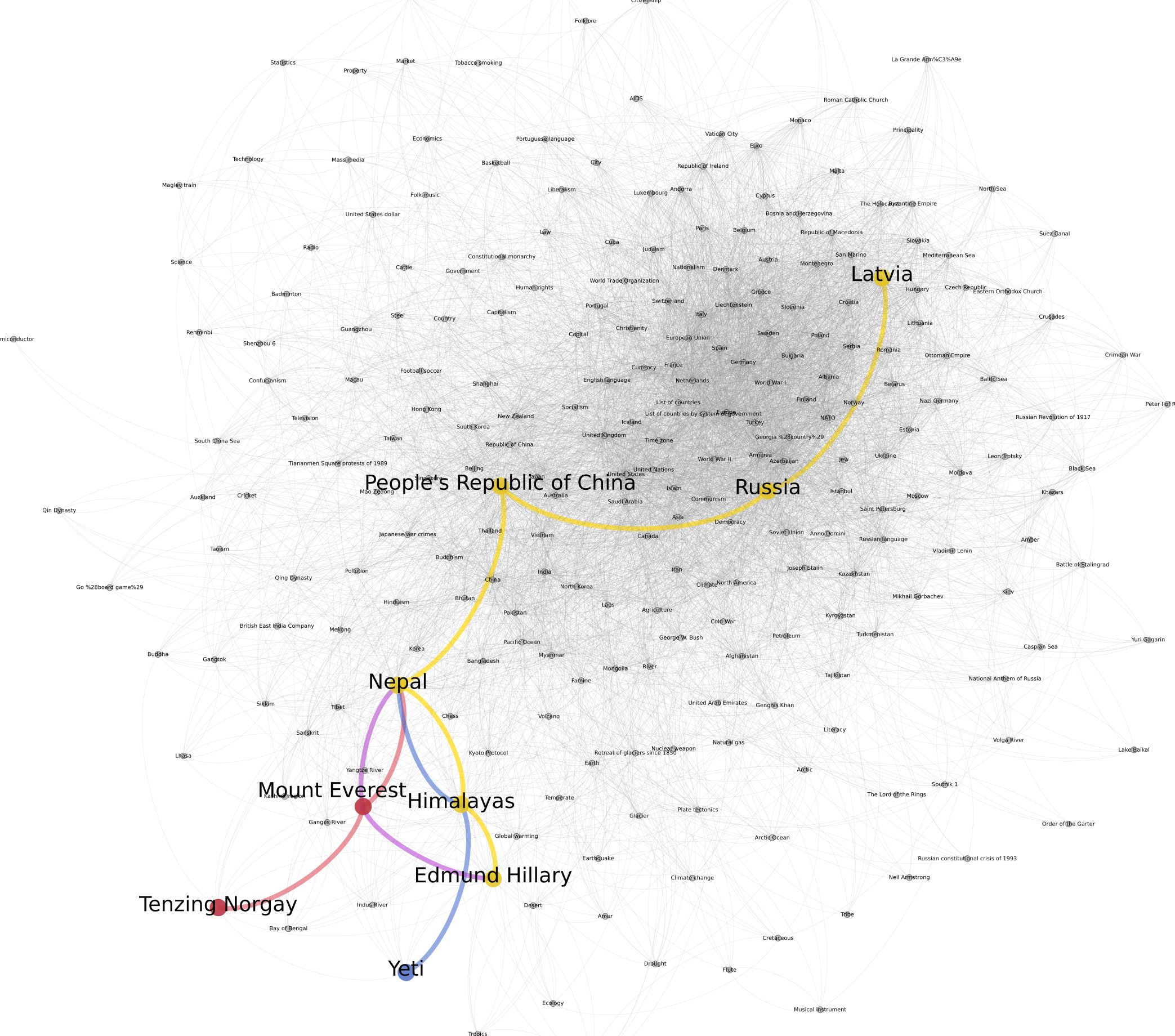}
\end{figure}

\begin{figure}[h]
    \centering
    \includegraphics[width=1.05\textwidth]{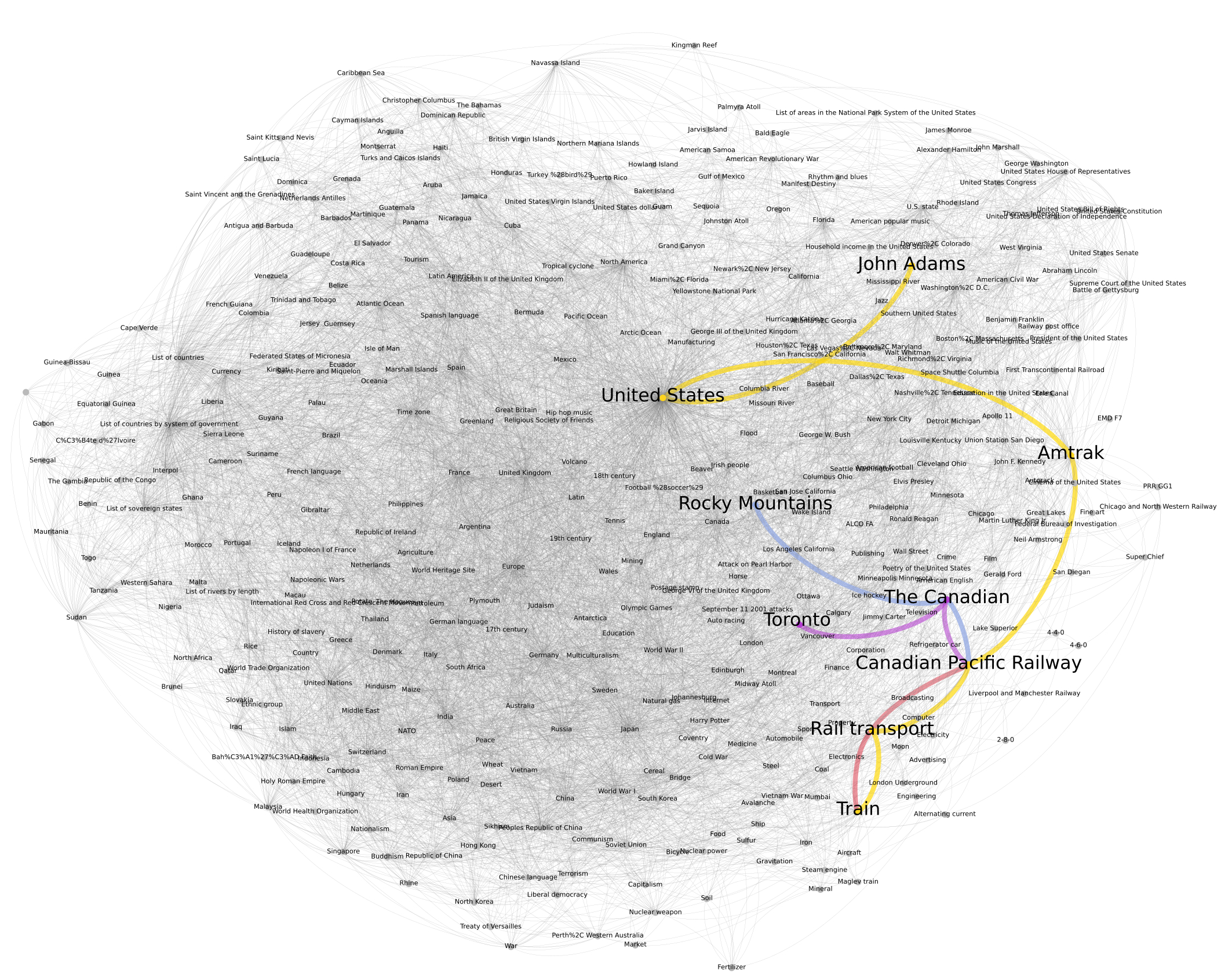}
\end{figure}

\end{document}